\documentclass{article}
\usepackage{spconf,amsmath,graphicx}

\title{TRADITIONAL METHOD INSPIRED DEEP NEURAL NETWORK FOR EDGE DETECTION}
\name{Jan Kristanto Wibisono and Hsueh-Ming Hang
			\thanks{The source code is available at https://github.com/jannctu/TIN}}
                   \address{Department of Electronics Engineering, National Chiao Tung University,Taiwan \\
                     jankristanto.03g@g2.nctu.edu.tw, hmhang@nctu.edu.tw}
\begin{document}
\ninept
\maketitle
\begin{abstract}
Recently, Deep-Neural-Network (DNN) based edge prediction is progressing fast. Although the DNN based schemes outperform the traditional edge detectors, they have much higher computational complexity. It could be that the DNN based edge detectors often adopt the neural net structures designed for high-level computer vision tasks, such as image segmentation and object recognition. Edge detection is a rather local and simple job, the over-complicated architecture and massive parameters may be unnecessary. Therefore, we propose a traditional method inspired framework to produce good edges with minimal complexity. We simplify the network architecture to include Feature Extractor, Enrichment, and Summarizer, which roughly correspond to gradient, low pass filter, and pixel connection in the traditional edge detection schemes. The proposed structure can effectively reduce the complexity and retain the edge prediction quality. Our TIN2 (Traditional Inspired Network) model has an accuracy higher than the recent BDCN2 (Bi-Directional Cascade Network) but with a smaller model.
\end{abstract}
\section{Introduction}
\label{sec:intro}
Edge detection is trying to find the high-contrast boundary that separates two regions. Edge detector is a fundamental operator in image processing. Along with the development of deep learning (DL), the development of Deep Neural Network (DNN) based edge detection also shows the advantage of high accuracy performance. On the BSDS500 dataset \cite{C1}, the traditional methods typically only achieve 0.59 ODS F-measure, but the DL-based methods can achieve 0.828 ODS \cite{C2}.
The current trend is to develop lighter edge detectors that maintain the detection accuracy. Fig. 1 shows both the detection accuracy and complexity (model size) of several well-known DNN based methods. Since the first deep edge detection was proposed, the ODS accuracy has increased from 0.782 (HED \cite{C3}) to 0.828 (BDCN \cite{C2}). Often, the DNN based edge detectors adopt the neural net structures designed for high-level computer vision tasks, such as image segmentation and object recognition. Therefore, they contain millions of parameters. Edge detection is a rather local and simple job; the over-complicated architecture and massive parameters may be unnecessary. For example, many researchers use VGGNet (Visual Geometry Group) \cite{C4} as their feature-based extractor, which was originally designed for image classification problem.   Consequently, their models have the capacity to extract a large number of features needed for image classification, but this large capacity could be overkill for edge detection purpose. 
\begin{figure}[htb]
\begin{minipage}[b]{1.0\linewidth}
  \centering
  \centerline{\includegraphics[width=8.5cm]{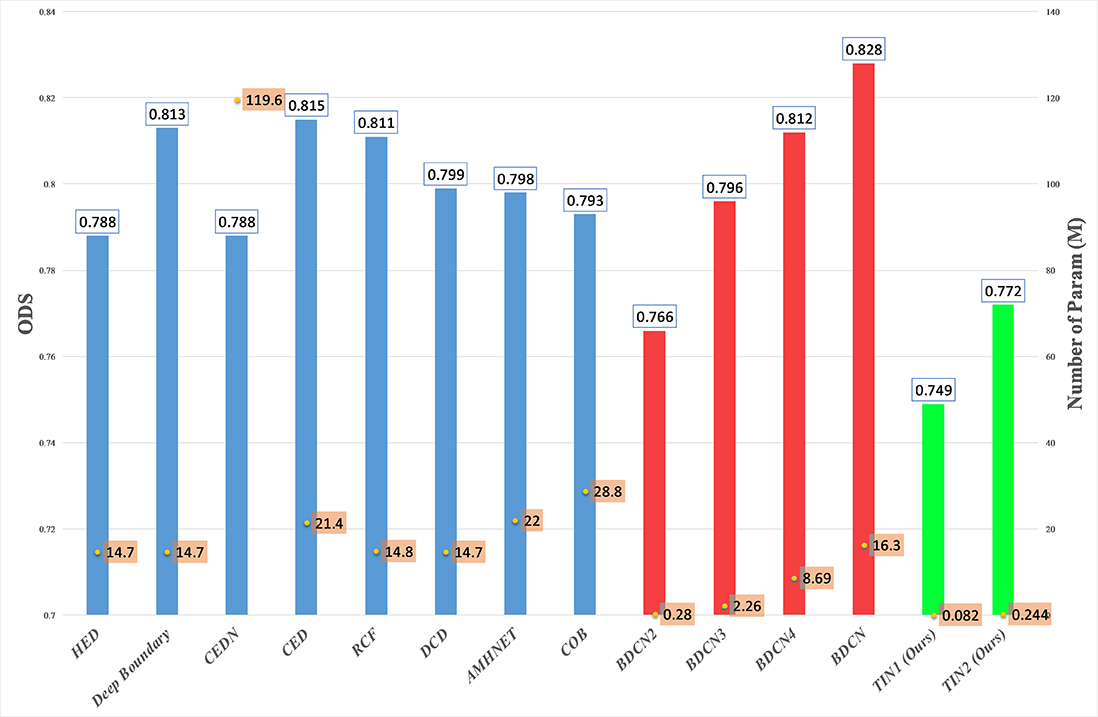}}
\end{minipage}
\caption{Comparison of complexity and accuracy performance among various edge detection schemes. Our proposed methods (Green). BDCN family (Red). Other methods (Blue). ODS (Transparent label). Number of Parameter (Orange label).}
\label{fig:res}
\end{figure}
In our opinion, edge detection is a much simpler job compared to image classification or semantic segmentation. In semantic segmentation, feature extractor is required to be able to recognize many different patterns of objects, whereas the edge detection only recognizes edge or non-edge. The traditional edge detectors \cite{C5,C6} typically only use gradients together with low-pass filtering and pixel connection.  This paper will focus on how to design an effective and lightweight deep learning framework to detect edges. We aim to reduce the complexity and retain the edge detection accuracy at the same time. In summary, in this paper, we propose a set of DNN based edge detectors which are inspired by the traditional methods. Our systems contain three basic modules: Feature Extractor, Enrichment, and Summarizer, which roughly correspond to gradient, low pass filter, and pixel connection in the traditional edge detection schemes. We compare the proposed method with the state-of-the-art methods. Our method has a much lower complexity at about the same edge detection accuracy.
\section{RELATED WORK}
\label{sec:format}
In the last few decades, many Edge Detection approaches have been proposed. We will highlight a few significant works. Sobel \cite{C5} and Canny \cite{C6} are the early works of Edge Detection. Sobel detection is calculating the magnitude of the gradient in an image using a set of 3x3 filters.
The next generation is the learning-based methods \cite{C7,C8,C9,C10,C11,C12,C13}. The learning-based methods often extract low-level features and use trained classifiers based on features to produce object boundaries. Even though these learning-based methods produce a much better performance compared to the conventional gradient methods, often their performance depends on how well the hand-crafted features are designed.
Recently, edge detection has been developed based on the deep convolutional neural networks. DeepEdge \cite{C14} and DeepCountour \cite{C15} are early learning-based edge detection schemes. Even though they were built on top of the deep neural network, they still adapted the notion of patches from the structured forest \cite{C12} and sketch tokens \cite{C11}. Different from the patches based on deep learning, HED \cite{C3} is an end-to-end fully convolutional neural network that accepts images as input and outputs the edge probability for every pixel. HED uses VGGNet \cite{C4} for the feature extraction. It fuses the all side-outputs of VGGNet features and minimizes the weighted cross-entropy loss function. HED is one of the most influential papers in the DNN-based edge detection. Then, Crisp Edge Detector (CED) \cite{C16} and Richer Convolutional Features (RCF) \cite{C17} were designed to enrich the use of the side-outputs. Also, there is an LPCB (Learning to Predict Crips Boundary) scheme \cite{C18}, which adjusts edge prediction by introducing a fusion loss function between the Dice coefficient \cite{C19} and the cross-entropy loss. The most recent work we found is BDCN \cite{C2}, which generates multiple ground-truth maps using different side-outputs. It is a family of schemes of different complexities and accuracies. It outperforms the other schemes at about the same complexities.
\section{PROPOSED METHOD}
\label{sec:pagestyle}
\subsection{Motivation}
\label{ssec:subhead}
A traditional edge detector often consists of gradient calculation, low-pass filter, pixel connection, and non-maximum suppression (NMS). Inspired by the traditional structure in building our system, our system consists of Feature Extractor, Enrichment, and Summarizer.
\subsection{Feature Extractor}
\label{ssec:subhead}
We start from a simple feature extractor, which is a 3x3 convolutional neural network layer. We mimic the gradient operators and after a few experiments, we settle on 16 feature channels and they are initialized with the 16 directional gradient kernels. The other CNN layers are initialized with zero-mean, 0.01 standard deviation Gaussian and zero biases. 
\begin{figure}[htb]
\begin{minipage}[b]{1.0\linewidth}
  \centering
  \centerline{\includegraphics[width=8.5cm]{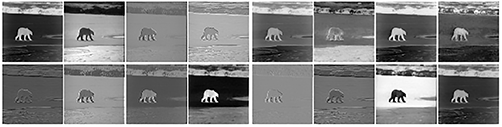}}
\end{minipage}
\caption{16 feature maps from the 1st Feature Extractor}
\label{fig:res}
\end{figure}
Although the feature extractor was designed to simulate the gradient operators, after training, some output feature maps show edges, while the other ones show, kind of, segmentation maps. Fig. 2 shows the 16 output maps from the first feature extractor. (Point A on Fig. 6) 
\subsection{Enrichment}
\label{ssec:subhead}
In a traditional edge detector, low-pass filters and the other more complicated schemes are often used to remove the noise or tiny/isolated edge candidates. We use dilated convolution\cite{C20} to build multiple scale filtering. The dilated convolution can capture a larger receptive field with fewer parameters. We call this module Enrichment because it also extracts object information in addition to edges. 
\begin{figure}[htb]
\begin{minipage}[b]{1.0\linewidth}
  \centering
  \centerline{\includegraphics[width=8.5cm]{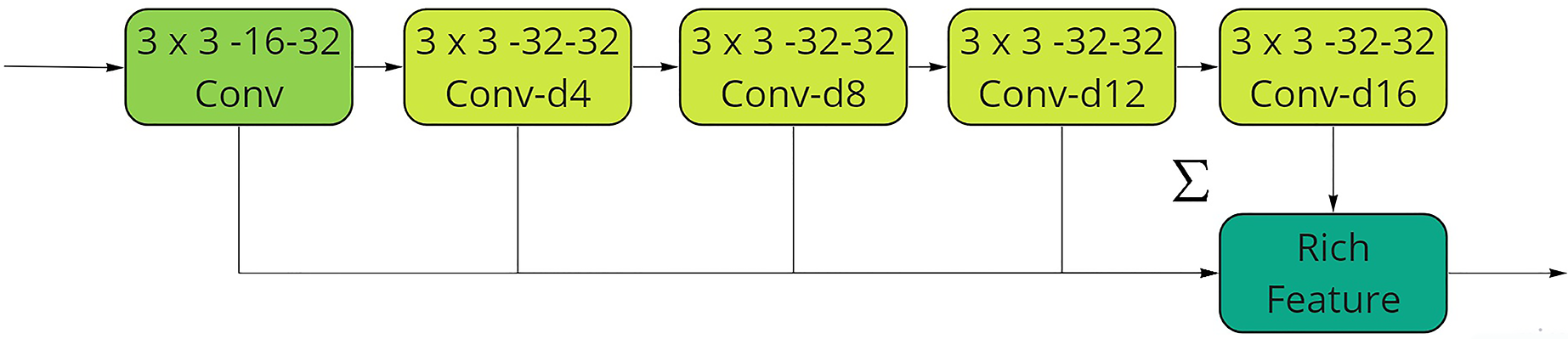}}
\end{minipage}
\caption{Enrichment component}
\label{fig:res}
\end{figure}
Fig. 3 shows the design of our Enrichment module. The 3x3 means the filter size is 3x3. The 16 means the number of input and the 32 means the number of output / number of filter. The outputs of each dilated filter are added together at the end.  Fig. 4 shows the 32 channel outputs of the Enrichment module (Point B on Fig. 6). The outputs are generally more blurred than their inputs (Feature Extractor outputs, Fig. 2). Some outputs contain only some portions of object edges; they seem to be complimentary to each other. 
\begin{figure}[htb]
\begin{minipage}[b]{1.0\linewidth}
 \centering
  \centerline{\includegraphics[width=8.5cm]{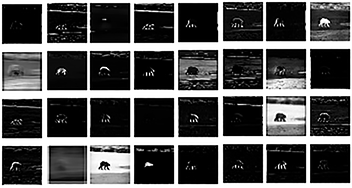}}
\end{minipage}
\caption{Intermediate outputs from Enrichment}
\label{fig:res}
\end{figure}
\subsection{Summarizer}
\label{ssec:subhead}
The last module tries to summarize the features generated by the Enrichment modules and to produce the final edges. We use eight 1x1 convolutional layers together with a sigmoid activation function.  Fig. 5(a) shows the 8 channel outputs of Summarizer 1 (Point C on Fig. 6), and (b) shows the final output of two summarizers (Point D on Fig. 6). 
\begin{figure}[htb]
\begin{minipage}[b]{1.0\linewidth}
  \centering
  \centerline{\includegraphics[width=8.5cm]{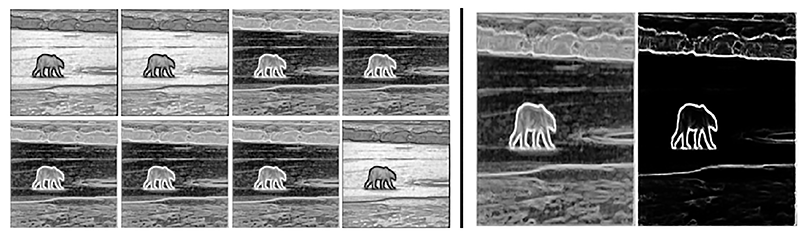}}
\end{minipage}
\caption{(a) Outputs of Summarizer 1 (b) Final Output}
\label{fig:res}
\end{figure}
\subsection{Loss Function}
\label{ssec:subhead}
Loss Function plays a critical role in our system. Because the edge pixels are much fewer than the non-edge pixels, we use a weighted cross-entropy loss \cite{C3}.  We compute the loss of each pixel prediction with respect to the ground-truth as follows. 
   \begin{equation}
    L(X_i;W)=
    \begin{cases}
      \alpha . log(1-P(X_i;W)), & \text{if}\ y_i=0 \\
      0, & \text{if}\ 0 <  y_i < th \\
      \beta . log(P(X_i;W)), & \text{otherwise}
    \end{cases}
  \end{equation}
where $y_i$, $X_i$,$W$ are the ground-truth, input, and weights, respectively. $P(X)$ is the sigmoid function, and $th$ is a threshold of value 64 in our case. By set the $th$, we ignore some weak edges because it is not significant and may confuse the network.  
 \begin{equation}
\alpha= \gamma . \frac{Y_+}{Y} \qquad \beta=\frac{Y_-}{Y}
  \end{equation}
$Y_+$ and $Y_-$ denote the numbers of edge and non-edge ground truth pixels, respectively. The parameter $\gamma$ is used to balance between edges and non-edges. Here, we use $y=1.1$ to emphasis more on edge.  This loss function is applied to all the side-outputs and the fused outputs. Thus the total loss function is Eq. 3, where K is the number of stages.
 \begin{equation}
    LL(W) = \sum_{k=1}^{K} \sum_{i=1}^{I} L(X_i^k;W) + \sum_{i=1}^{I} L(X_i^{fuse};W)
  \end{equation}
\subsection{Post-processing}
\label{ssec:subhead}
Similar to \cite{C2,C3,C16,C17,C18}, we use multi-scale testing and non-maximum suppression (NMS) as our post-processing. First, we resize the input image to three different resolutions, 0.5x,1x, and 1.5x, then feed them into the network. We resize the outputs into the original size and average them to produce the final result. The purpose of NMS is to thin the edge prediction maps.
\subsection{Traditional Method Inspired Edge Detectors}
\label{ssec:subhead}
We propose two architectures using the components discussed earlier. We called them Traditional Inspired Network 1 (TIN 1) and Traditional Inspired Network 2 (TIN 2). TIN2 is simply a stack of two TIN1, but the stage one feature map is downsampled (max-pooling) in half before entering the second stage. Fig. 6 shows the architecture of TIN1. Fig. 7 is TIN2. Our input is the original color image without any preprocessing. The input is passed to the Feature Extractor 1, which is 16 filters of 3x3 convolution as we described before. These features go to the Feature Extractor 2 with random initialization. Each Feature Extractor passes its outputs to the Enrichment and Summarizer modules. At the end, we add the Summarizer output features and fuse them by a 1x1 convolution.
\begin{figure}[htb]
\begin{minipage}[b]{1.0\linewidth}
  \centering
  \centerline{\includegraphics[width=8.5cm]{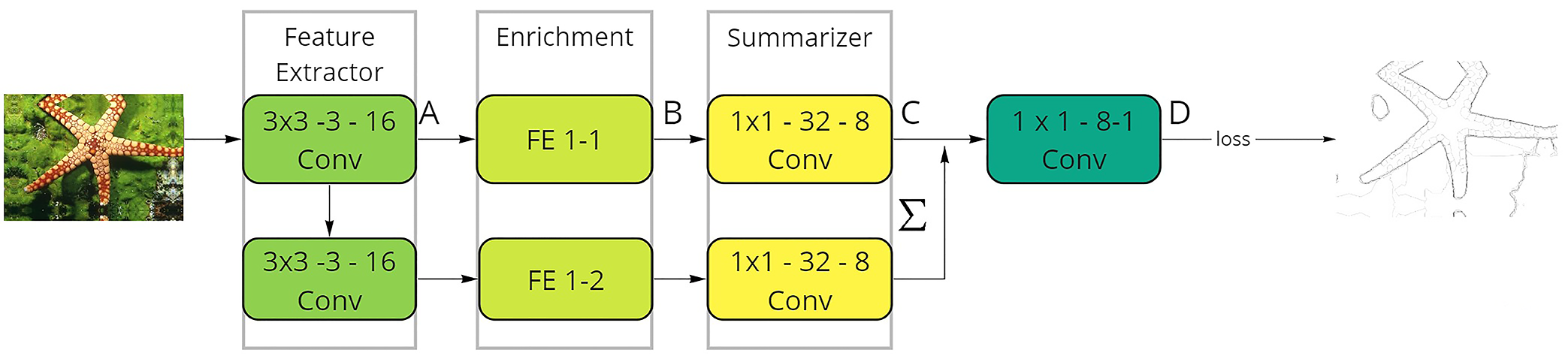}}
\end{minipage}
\caption{TIN1 Architecture}
\label{fig:res}
\end{figure}
In training, the loss function is optimized to tune the parameters. The TIN2 (Fig.7) Feature Extractor has 64 filters. TIN2 produces multiple prediction outputs. To fuse the outputs of both stages, we apply the bilinear interpolation to the second stage outputs. Concatenating stage 1 outputs and the interpolated stage 2 outputs, we form the final edge map using another 1x1 convolution. The bilinear interpolation may be replaced by a deconvolutional layer but it produces similar results in our experiment.
\begin{figure}[htb]
\begin{minipage}[b]{1.0\linewidth}
  \centering
  \centerline{\includegraphics[width=8.5cm]{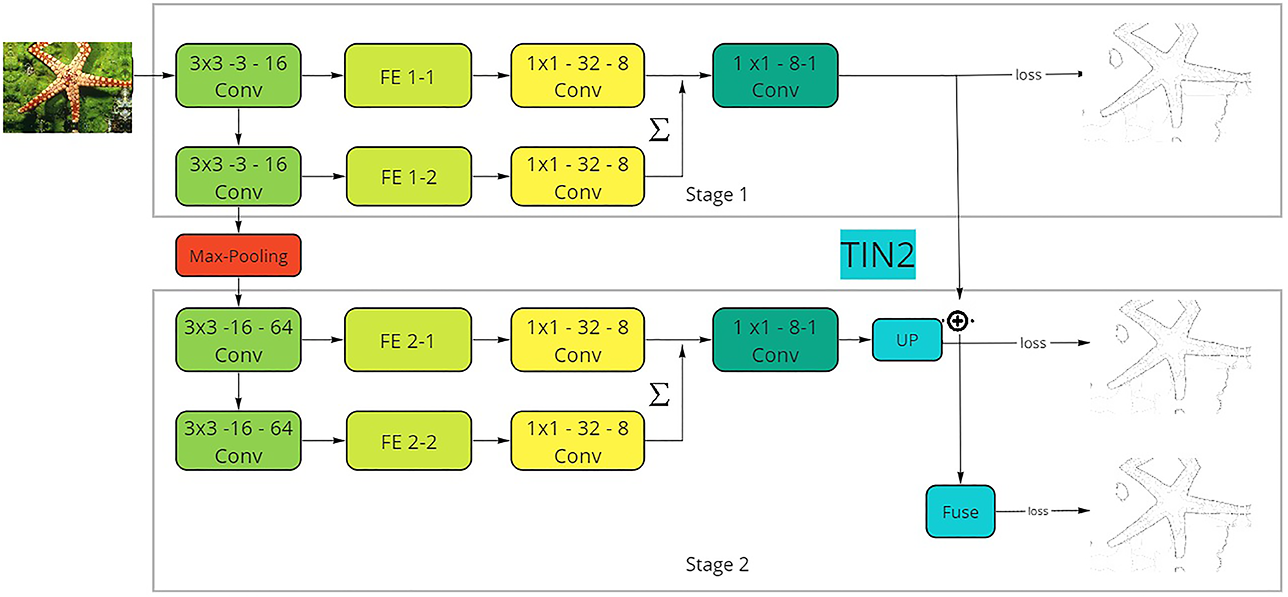}}
\end{minipage}
\caption{TIN2 Architecture}
\label{fig:res}
\end{figure}
\section{EXPERIMENTS}
\label{sec:typestyle}
\subsection{Implementation Details }
\label{ssec:subhead}
Our network is implemented on Pytorch. Except for the Feature Extractor 1, the model (filter weights) are initialized by using the Gaussian distribution with zero-mean and standard deviation 0.01, and the biases are initialized with 0. The hyper-parameters are as follows: mini-batch size=1, learning rate=(1e-2), weight decay=(5e-4), momentum=0.9, and training epochs=120. We divide the learning rate by 10 for every 10 epochs. The SGD solver is used for optimization.  All the experiments are conducted on NVIDIA GeForce 2080Ti GPU with 11G memory. The maximum tolerance is 0.0075 for BSDS500 and 0.011 for the NYUDv2 dataset, as defined in the previous works  \cite{C2,C3,C17}.
\subsection{Datasets }
\label{ssec:subhead}
We train our method using BSDS500 \cite{C1}, PASCAL VOC \cite{C21}, and NYUDv2 \cite{C22} and evaluate it using BSDS500 and NYUDv2, which has been used by many edge detection researches \cite{C2,C3,C16,C17,C18}. Data augmentation has the ability to increase the amount of data available for training the model without actually collecting new data. We implement the data augmentation strategy in \cite{C2,C3,C17} by rotating images to 16 different angles, flipping them, and also scaling them.
\subsection{Evaluation Metrics }
\label{ssec:subhead}
We evaluate our schemes using the popular F-measure \cite{C1}. The output of our network is an edge probability map; we need a threshold to produce an edge image. There are two options to set this threshold. First, we set a threshold for all images in the dataset, which is called Optimal Dataset Scale (ODS) \cite{C1}. And the second is Optimal Scale Image (OIS) \cite{C1}, which chooses one threshold for each image.       
\subsection{Experimental Result and Comparison }
\label{ssec:subhead}
We compare our proposed method with several state-of-the-art methods. We categorize the comparison into three. First, a comparison against the traditional approaches. Second, comparison against the machine learning-based approaches, and the last comparison against the deep learning approach. For this comparison, our architecture is trained on the BSDS500 dataset and PASCAL VOC dataset. We evaluate the model using the BSDS500 test set similar to \cite{C2,C3,C14,C15,C16,C17,C18,C24}.  
\begin{figure}[htb]
\begin{minipage}[b]{1.0\linewidth}
  \centering
  \centerline{\includegraphics[width=8.5cm]{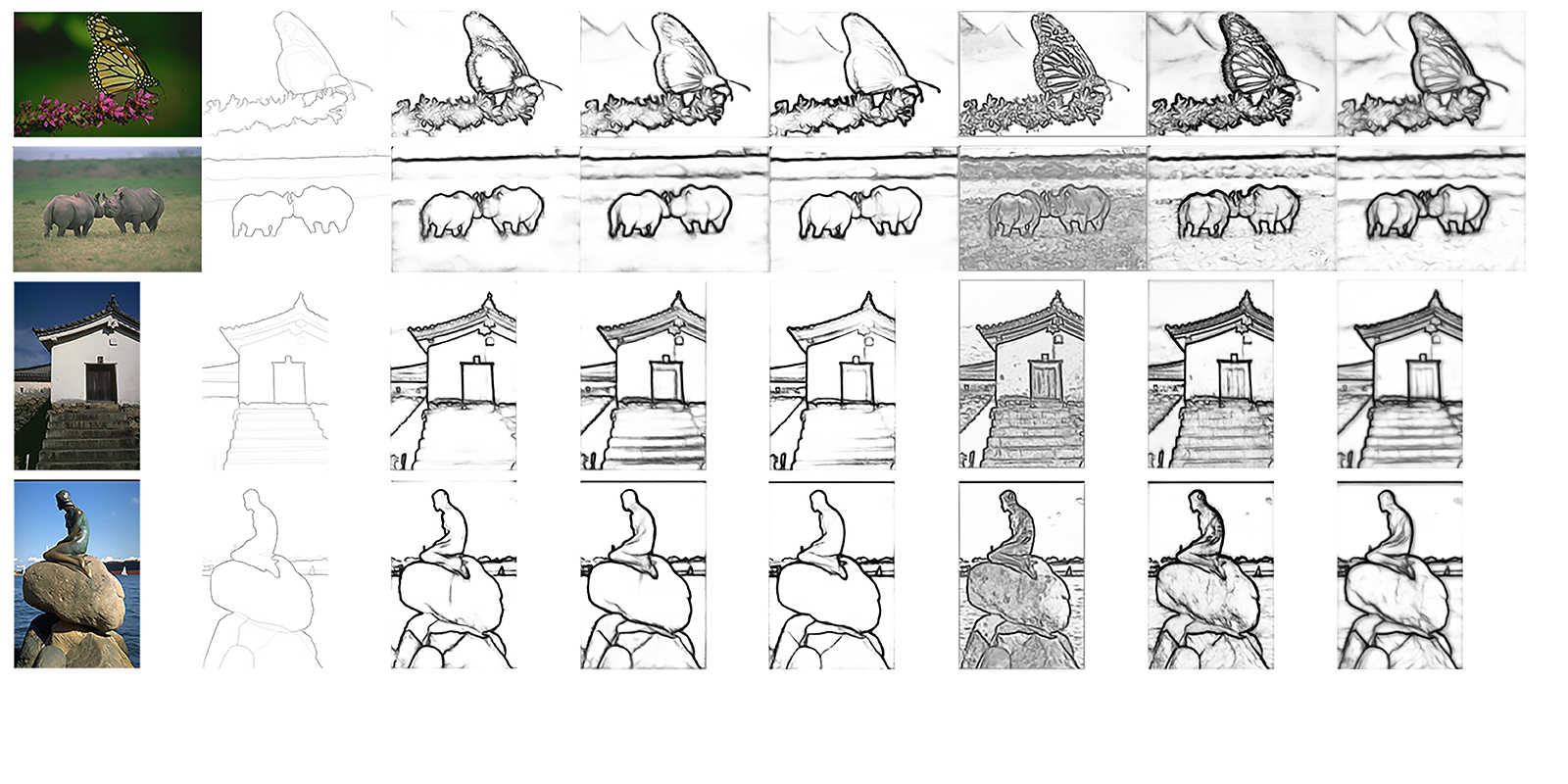}}
\end{minipage}
\caption{Edge Detection Comparison on BSDS500. From left to right: input, ground truth, HED \cite{C3}, RCF\cite{C17}, BDCN5\cite{C2}, BDCN2\cite{C2}, TIN1(Ours), TIN2(Ours) }
\label{fig:res}
\end{figure}
Table 1 shows the quantitative results against the other edge detection approaches. As we can see, our TIN1 has an extremely small model, at around 80K parameters, which outperforms the traditional methods and most machine learning methods. TIN2 has 0.24M parameters but it achieves an ODS of 0.772, outperforms the BDCN2 \cite{C2}, which has 0.28M parameters. Fig. 8 shows visual comparisons of our approach and a few others.
\begin{table}[h!]
  \begin{center}
    \caption{Comparison on BSDS500 test dataset}
    \label{tab:table1}
    \begin{tabular}{c c c c c} 
      \textbf{Method} & \textbf{ODS} & \textbf{IOS} & \textbf{No Params} & \textbf{FPS}\\
      \hline
      Canny \cite{C6} & 0.611 & 0.676 &  & 28 \\
      \hline
	  gPb-UCM \cite{C1} &	0.729 & 0.755 &	 & 1/240\\
	  SCG\cite{C10}  &	0.739  & 0.758  &  & 1/18\\
      SF\cite{C12} & 0.743 &	0.763 &	 & 2.5\\
      OEF \cite{C13} &	0.749 &	0.772 &	 & 2/3\\
      \hline
	 DeepEdge \cite{C14} &	0.753 &	0.772 &	- &	1/1000\\
DeepContour\cite{C17} &	0.757 &	0.776 &	- &	1/30\\
HED \cite{C3} &	0.782 &	0.804 &	14.7 &	30 \\
CEDN \cite{C4}  & 0.788  & 0.804  & 119.6  &	10\\
RCF \cite{C17}  &	0.806 &	0.823 &	14.8 &	30\\
RCF-MS \cite{C17} &	0.811 &	0.830 &	14.8 &	10\\
CED \cite{C16} &	0.794 &	0.811 &	21.4 &	30\\
CED-MS \cite{C16} &	0.815 &	0.834 &	21.4 &	10\\
LPCB \cite{C18} &	0.808 &	0.824 &	- &	30\\
LPCB-MS \cite{C18}	 &0.815 &	0.834 &	- &	10\\
BDCN2 \cite{C2}&	0.766&	-&	0.28	& \\
BDCN3 \cite{C2}	&0.796&	-&	2.26	& \\
BDCN4 \cite{C2}&	0.812&	- &	8.69	& \\
BDCN \cite{C2}&	0.820&	0.838&	16.3	& \\
BDCN-MS \cite{C2}&	0.828&	0.844&	16.3	& \\
     \hline
TIN1 (ours)&	0.749&	0.772&	0.08&	30\\
TIN2 (ours)&	0.772&	0.795&	0.24&	30\\
    \end{tabular}
  \end{center}
\end{table}
We also train our architectures on the NYUDv2 \cite{C22} training set and validation set. We use Gupta et al. \cite{C23} version that split the data set into 381 training, 414 validation and 654 testing images set. We train the NYUDv2 models using the same parameters for BSDS500. Since the NYUDv2 dataset has much fewer data, compared to BSDS500, we increase the epochs to archive a similar number of iterations. 
\begin{table}[h!]
  \begin{center}
    \caption{Comparison on BSDS500 test dataset}
    \label{tab:table1}
    \begin{tabular}{c c c c} 
      \textbf{Method} & \textbf{Model} & \textbf{ODS} & \textbf{IOS} \\
      \hline
      gPb-UCM \cite{C1}	  &RGB &	0.632 &	0.661\\
OEF \cite{C13} &RGB &	0.651 &	0.667\\
SE \cite{C12} &RGB &		0.695 &	0.708\\
SE+NG+\cite{C23} &RGB &		0.706 &	0.734\\
      \hline
HED \cite{C3} &	RGB &	0.720 &	0.734\\
HED \cite{C3} &	HHA &	0.682 &	0.695\\
HED \cite{C3}	 &RGB+HHA &	0.746 &	0.761\\
      \hline
RCF \cite{C17}	 &RGB &	0.729 &	0.742\\
RCF \cite{C17}	 &HHA &	0.705 &	0.715\\
RCF \cite{C17}	 &RGB+HHA &	0.757 &	0.771\\
      \hline
BDCN5 \cite{C2}	 &RGB &	0.748 &	0.763\\
BDCN5 \cite{C2}	 &HHA &	0.707 &	0.719\\
BDCN5 \cite{C2}	 &RGB+HHA &	0.765 &	0.781\\
      \hline
TIN1 (Ours)	 &RGB	 &0.706 &	0.723\\
TIN1 (Ours) &	HHA &	0.661 &	0.681\\
TIN1 (Ours)	 &RGB+HHA &	0.729 &	0.750\\
      \hline
TIN2 (Ours)	 &RGB &	0.729 &	0.745\\
TIN2 (Ours) &	HHA &	0.705 &	0.722\\
TIN2 (Ours)	 &RGB+HHA &	0.753 &	0.773\\
    \end{tabular}
  \end{center}
\end{table}
For each architecture, TIN1, and TIN2, we train two models: RGB model and HHA representation model. In the evaluation stage, we increase the maximum tolerance to 0.011 as used in \cite{C2,C3,C17}. We compare our performance with some other approaches, such as gPb-UCM \cite{C1}, Structured Forest \cite{C12}, HED \cite{C3}, RCF \cite{C17}, and BDCN \cite{C2}. Our models achieve rather good performance compared to the other complicated models. The numerical metrics are shown on Table 2. 
\section{CONCLUSIONS}
\label{sec:typestyle}
In this paper, we propose a traditional method inspired DNN architecture for edge detection. Different from the previous approaches that borrow the design from the other DNN architectures designed for the other purposes, we develop our system based on the concepts of classical edge detection. We carefully build our model step-by-step and the final system can achieve good edge quality with a much simpler model. Our experiments show that our TIN1 produces good results with an extremely small model. Our TIN2 model has an accuracy higher than the recent BDCN2 but with a smaller model.
\section{ACKNOWLEDGEMENT}
\label{sec:foot}
This work is partially supported by the Ministry of Science and Technology, Taiwan under Grant MOST 109-2634-F-009-020 through Pervasive AI Research (PAIR) Labs, National Chiao Tung University, Taiwan.

\newpage
\bibliographystyle{IEEEbib}
\bibliography{strings,refs}
\end{document}